%% file: main.tex
\documentclass[12pt]{iopart}

\usepackage{graphicx}
\usepackage{makecell}
\usepackage{xcolor}
\usepackage{soul}
\expandafter\let\csname equation*\endcsname\relax
\expandafter\let\csname endequation*\endcsname\relax
\usepackage{amsmath}
\usepackage[noadjust]{cite}
\usepackage{lineno}
\usepackage[normalem]{ulem}
\usepackage{url}

\begin{document}

\title[Bioinspired Smooth Neuromorphic Control for Robotic Arms]{Bioinspired Smooth Neuromorphic Control for Robotic Arms}

\author{Ioannis Polykretis$^{1, 2}$ \footnote{IP was partially funded by the Onassis Foundation scholarship}, Lazar Supic$^{1}$, and Andreea Danielescu$^{1}$}

\address{$^{1}$ Accenture Labs, San Francisco, CA, USA}
\address{$^{2}$ Computer Science Department, Rutgers University, New Brunswick, NJ, USA}

\ead{ioannis.polykretis@accenture.com}
\vspace{10pt}

\begin{abstract}
\input{tex_input/abstract.tex}

\end{abstract}

%
%
%
%
%

\section{Introduction}
\input{tex_input/introduction_new.tex}

\section{Related Work}
\input{tex_input/related_work.tex}

\section{Biological Background}
\input{tex_input/bio_background}

\section{Methods}
\input{tex_input/methods.tex}
  
\section{Results}
\input{tex_input/results.tex}

\section{Discussion}
\input{tex_input/discussion.tex}


\section{Data Availability Statement}
The data that support the findings of this study are available upon request from the authors.

\section*{References}
\bibliographystyle{abbrv}
\bibliography{myref}

\end{document}

%% file: tex_input/abstract.tex
Beyond providing accurate movements, achieving smooth motion trajectories is a long-standing goal of robotics control theory for arms aiming to replicate natural human movements. Drawing inspiration from biological agents, whose reaching control networks effortlessly give rise to smooth and precise movements, can simplify these control objectives for robot arms. Neuromorphic processors, which mimic the brain's computational principles, are an ideal platform to approximate the accuracy and smoothness of biological controllers while maximizing their energy efficiency and robustness. However, the incompatibility of conventional control methods with neuromorphic hardware limits the computational efficiency and explainability of their existing adaptations. In contrast, the neuronal subnetworks underlying smooth and accurate reaching movements are effective, minimal, and inherently compatible with neuromorphic hardware. In this work, we emulate these networks with a biologically realistic spiking neural network for motor control on neuromorphic hardware. The proposed controller incorporates experimentally-identified short-term synaptic plasticity and specialized neurons that regulate sensory feedback gain to provide smooth and accurate joint control across a wide motion range. Concurrently, it preserves the minimal complexity of its biological counterpart and is directly deployable on Intel's neuromorphic processor. Using the joint controller as a building block and inspired by joint coordination in human arms, we scaled up this approach to control real-world robot arms. The trajectories and smooth, bell-shaped velocity profiles of the resulting motions resembled those of humans, verifying the biological relevance of the controller. Notably, the method achieved state-of-the-art control performance while decreasing the motion jerk by 19\% to improve motion smoothness. Overall, this work suggests that control solutions inspired by experimentally identified neuronal architectures can provide effective, neuromorphic-controlled robots.

%% file: tex_input/introduction_new.tex
Modern robot arms perform a wide variety of complex and repetitive tasks, both in isolated industrial environments~\cite{brogaardh2007present} and in close interaction with humans~\cite{ajoudani2018progress}. The efficiency of the arms' performance requires the determination of the best trajectory for the execution of the task~\cite{gasparetto2015path, ata2007optimal}. This trajectory can be determined by optimizing several variables. Beyond the paramount objective of control precision, a crucial optimization variable is the smoothness of the end-effector's trajectory~\cite{constantinescu2000smooth}. Control smoothness not only improves position control~\cite{wang2019smooth, liu2016rapid} but also slows down motor wear~\cite{rivera2010extending} by avoiding abrupt load changes and ensures a safe and comfortable human-robot interaction by avoiding damage to the robot, its user, or their environment~\cite{hochberg2006neuronal}. Patient-assistive robotic arms and other human-robot collaboration contexts especially benefit from smooth robotic control as they require low end-effector velocity when the arms enter a safety bubble surrounding the user~\cite{ehrlich2022adaptive}. For this, numerous methods aim to optimize control smoothness by introducing kinematic constraints to the motion velocity, acceleration, and jerk~\cite{xiao2012smooth, liu2016rapid, wang2019smooth}. However, these constraints further complicate the already challenging optimization problem.    

Humans, however, demonstrate smooth trajectories when performing reaching movements~\cite{uno1989formation} as early as their first postnatal months~\cite{sacrey2010development}. Primates and rodents also achieve similar reaching behaviors during goal-oriented movements~\cite{seki2003sensory, fink2014presynaptic}. The resulting trajectories satisfy the optimality criteria discussed above~\cite{flash1985coordination, whishaw1996endpoint, todorov2004optimality} seemingly effortlessly and in the presence of limited energy and computational resources. The motion smoothness depends on specialized neuronal microcircuits that form recurrent loops and refine sensory feedback~\cite{fink2014presynaptic}. Inhibitory interneurons critically enrich the computational repertoire of these microcircuits by affecting not only other neurons but also the dynamics of the synapses that interconnect them. The specialized synapse-modulating inhibitory neurons target presynaptic axons and modify their neurotransmitter release probability~\cite{wu1997presynaptic}. The effect of inhibition at this presynaptic locus has a divisive effect on feedback signals and provides a versatile mechanism of sensory gain control~\cite{rossignol2006dynamic}. This dynamic feedback gain provides accurate and stable control across diverse motion ranges in biological agents. Including such neuroscientific principles in robot control could potentially give rise to novel solutions that would be as effective as their conventional counterparts and computationally efficient, adaptable to a wide range of stimuli, and explainable in terms of behavior and malfunction~\cite{polykretis2020astrocyte, polykretis2022spiking, kreiser2020chip}.

Neuromorphic algorithms and hardware platforms~\cite{davies2018loihi, furber2014spinnaker, moradi2017scalable, pehle2022brainscales} also draw inspiration from the same neuroscientific principles, attempting to emulate computation in the brain~\cite{calimera2013human, thakur2018large}. Consequently, this computational framework is ideal for integrating biological inspiration into robot control. Neuromorphic solutions already provide additional benefits, such as low latency, low power consumption, robustness, and potentially higher accuracy, which have been proven primarily outside of robotic applications~\cite{taunyazov20event, imam2020rapid}. These benefits are critical for all robotic applications, but are especially critical for mobile real-time human-robotic applications, which have additional power constraints. Brain-inspired neuromorphic methods have successfully targeted a wide range of tasks: from dynamic navigation and mapping~\cite{milde2017obstacle, tang2019spiking} to trajectory generation~\cite{michaelis2020robust} and joint control~\cite{jimenez2012neuro, stagsted2020event, zaidel2021neuromorphic}. On a broader perspective, biological inspiration has provided valuable insights for the whole spectrum of robot control~\cite{winters2012biomechanics}: Full-scale behaviors such as locomotion~\cite{strohmer2020flexible, ijspeert2008central, polykretis2020astrocyte} and swimming~\cite{crespi2005amphibot, ijspeert2007swimming} benefit from the knowledge of Central Pattern Generators~\cite{grillner1998intrinsic}; Low-level joint control exploits neuroscientific knowledge of neuromotor control models to design stable and robust variants of the classical PID controller~\cite{yin2003motion, tan2011stable}. However, existing methods, be they neuromorphic or merely brain-inspired, have yet to emulate the characterized neuronal microcircuits that drive smooth movements in animals to explore the optimization of the trajectory smoothness in robotic control.

\begin{figure}[!t]
\vspace{-5.2pt}
\centering
\includegraphics[scale=1.3]{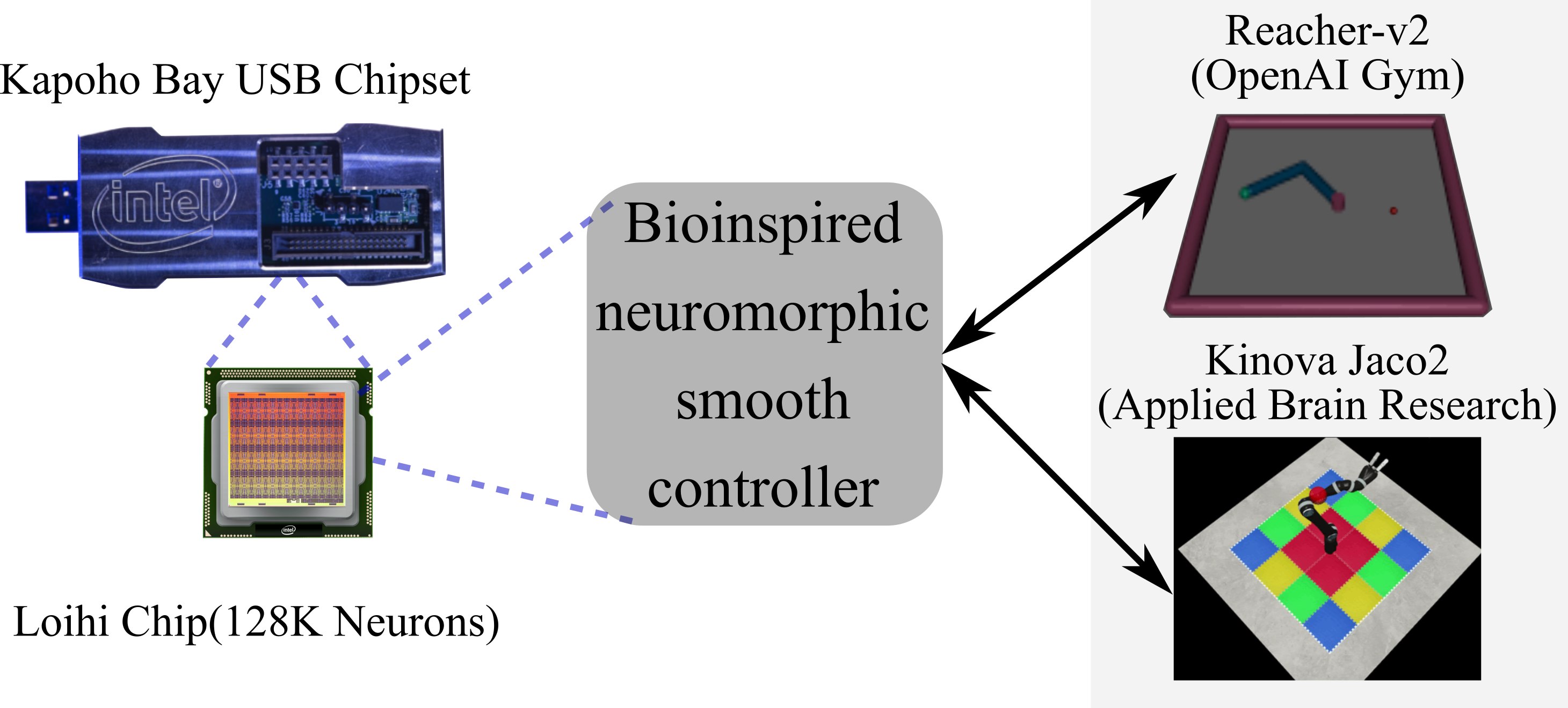}
\caption{Overview. An SNN-based controller inspired by the biological neuronal networks that control reaching movements provided smooth and accurate control of robot arms. The network was tested on a CPU and deployed on Intel's Loihi neuromorphic chip to drive the end-effector of two increasingly complex simulated robot arms. The resulted control compared favorably with a state-of-the-art PID, in both smoothness and accuracy.}
\label{overview}
\vspace{-5.2pt}
\end{figure}

This work addresses this gap by developing a compact SNN architecture that allows for smooth and accurate neuromorphic control. The proposed SNN architecture attempts to faithfully emulate the function of the specialized neuronal microcircuits that provide smooth joint control in primates and rodents~\cite{fink2014presynaptic}. It comprises antagonistic extensor and flexor motor neurons that modify the controlled variable and sensory proprioceptor ($PPC$) neurons that ensure smoothness by regulating the sensory feedback to the motor neurons. This biologically plausible architecture is directly deployable on Intel's neuromorphic research processor, the Loihi. The method is evaluated in the control of two increasingly complex robotic arms: i) a two-link arm (Reacher-v2) and ii) a 6-DOF real-world robotic arm (Jaco). The controller matches the performance of a conventional PID controller when evaluated on standard controller metrics (rise time, settling time, and overshoot)~\cite{levine2018control} while also improving control smoothness. The combination of smooth control for each joint with a biologically inspired coupling of the joints gives rise to end-effector trajectories that resemble those of human arms. Therefore, this work demonstrates that drawing inspiration from experimentally identified neuronal architectures can successfully emulate their function in neuromorphic-controlled robots.

%% file: tex_input/related_work.tex
Conventional control methods have matured over several decades and are now very effective for controlling joints of multi-link robotic arms. The theoretical underpinnings of these methods rely on systems of nonlinear differential equations or dynamical systems~\cite{luo1985lq, malki1997fuzzy}. The continuous, high-precision and synchronized computations that these methods rely on hinder their compatibility with modern neuromorphic processors. This incompatibility prevents them from exploiting the energy efficiency, low latency, and robustness provided by neuromorphic computing \cite{dewolf2023neuromorphic}.

In an effort to combine the performance of conventional methods with the benefits of neuromorphic computing, recent solutions have attempted to translate PID controllers on both analog~\cite{jimenez2012neuro} and digital~\cite{stagsted2020event, stagsted2020towards} neuromorphic processors. Although these approaches achieve competitive control performance, they suffer in terms of computational efficiency and interpretability. These come as a necessary sacrifice when the precise, continuous values of conventional control methods need to be represented and used in the low-precision, discrete and asynchronous computations performed on neuromorphic hardware. Specifically, these approaches represent continuous values utilizing large neuronal populations, whose size scales as a function of the required control precision to allow for distinct value representation~\cite{zaidel2021neuromorphic, stagsted2020event, stagsted2020towards}. After encoding the continuous quantities, these methods perform the necessary computations by exploiting large neuronal arrays, further increasing the computational burden~\cite{zhao2020closed, stagsted2020event, glatz2019adaptive}. The large number of required components introduces significant parameter tuning, which is inefficient or even infeasible in hardware such as mixed signal neuromorphic processors with inherent inaccuracies~\cite{glatz2019adaptive}. Lastly, the resulting architectures are platform-specific~\cite{jimenez2012neuro}, requiring a complete rethinking of their design to translate to different processors.

This work addresses some of these drawbacks by proposing a biologically inspired SNN for the smooth control of robotic joints. Our architecture draws inspiration from the experimentally identified neuronal connectivity that gives smooth reaching movements in rodents~\cite{fink2014presynaptic} and primates~\cite{seki2003sensory}. Mimicking the connectivity of its biological counterpart, our controller architecture is not only minimal in terms of computational resources but is also interpretable. Moreover, the core of our architecture can be deployed on any neuromorphic platform that includes spiking compartments with modifiable connectivity, requiring only minor design changes to implement the additional adaptation mechanisms (synaptic facilitation and presynaptic inhibition).

%% file: tex_input/bio_background.tex
To replicate the smoothness of reaching movements in biological agents with robot arms, we must first understand the key factors that contribute to it. As smoothness is related to movement continuality~\cite{balasubramanian2015analysis}, the discrete endpoints of reaching movements may  introduce discontinuity when combined with the inertial properties of the limbs. Therefore, the two sensory-dependent movement components, the initiation and termination, are crucial for the smoothness.

Movement initiation must be sensitive to subtle sensory cues to ensure responsiveness, but it also needs to be gradual to ensure continuity. One of the mechanisms that contributes to the satisfaction of these constraints is short-term synaptic plasticity~\cite{tsodyks1998neural}. Specifically, synaptic facilitation, i.e., the input-dependent, temporary increase of synaptic weights, allows for temporally separated presynaptic inputs, which would otherwise have no effects, to accumulate and induce postsynaptic spikes. This accumulation becomes crucial for motor neurons, which receive presynaptic stimuli that encode external (sensory) and internal (proprioceptive) error representations. In rate-encoding representations, errors of small magnitude correspond to low spiking frequencies. Therefore, the accumulation mediated by synaptic facilitation can activate motor neurons even at low presynaptic firing rates~\cite{nadim2000role}, and allows for the correction of small deviations, ensuring accurate control. Concurrently, this progressive transfer of presynaptic, error-encoding activity to motor neurons ensures gradual movement initiation.

Effective movement termination depends on finely regulated sensory feedback. Without that, the motor circuits, which are inherently prone to oscillation~\cite{baker2007oscillatory}, would fail to achieve smooth limb trajectories. Extensive evidence highlights the role of inhibitory neurons in regulating sensory gain~\cite{windhorst1996role, schlaghecken2002motor}, with the most prominent inhibitory neurotransmitter, gamma-aminobutyric acid (GABA), being a key factor~\cite{stagg2011role}. Whereas the majority of inhibitory neurons directly target motor or premotor neurons~\cite{arber2012motor}, a small subset of GABA-ergic neurons target the axons of presynaptic sensory afferents and reduce their neurotransmitter release probability~\cite{wu1997presynaptic}. While postsynaptic inhibition is far less effective in gain scaling due to its subtractive nature~\cite{capaday1987method, capaday1987difference}, the divisive effect of presynaptic inhibition is a versatile mechanism of sensory gain control~\cite{rossignol2006dynamic} with experimental evidence suggesting that GABA-mediated presynaptic inhibition regulates the sensory gain in primates and rodents~\cite{fink2014presynaptic, seki2003sensory} to suppress forelimb oscillation and provide smooth reaching movements. The ablation of this GABA-ergic presynaptic inhibition highlights its importance since it results in damped
harmonic oscillations~\cite{fink2014presynaptic}.

The architecture proposed in this paper emulates these two crucial mechanisms to achieve smooth end-effector trajectories. First, synaptic facilitation is implemented as a local plasticity mechanism. The synaptic weights temporarily increase in response to presynaptic activity to model how synaptic vesicles get gradually entrained in the facilitatory synapses~\cite{tsodyks1998neural}. Then, the mechanism of presynaptic inhibition is emulated as a separate disynaptic pathway. There, a presynaptic inhibitory neuron (PSI) resembles the effects of the GABA-ergic neurons on the sensory afferent terminals and induces a divisive effect on the depolarization of the motor neurons~\cite{fink2014presynaptic}. The implementation details are described in Section \ref{sec:wgt_adapt}. \\

%% file: tex_input/methods.tex
The objective of the SNN controller is two-fold: i) to ensure control accuracy by decreasing the control error, i.e. the difference between the current control variable value $\theta$ and the desired control variable value, $\theta^d$, and ii) to ensure smooth control through a gradual initial increase of the control variable and a timely steady decline to avoid overshooting the target. The architecture of the controller is shown in Fig. \ref{architecture}. In the following sections, we first provide a detailed description of the components of the single-joint controller and then, we describe the neuromorphic deployment of the network. Finally, we explain the addition of joint coordination to support multi-DOF arms.

\subsection{Architecture of the 1-DOF SNN controller}
\subsubsection{Input and Feedback Encoding} 
The proposed SNN controls a single variable by comparing its actual value $\theta$, which is internally fed back, with the desired value $\theta^d$, which is the only external input provided to the controller. The values of these two quantities are encoded into currents by the current converter modules (CCMs) and then stimulate the neurons of the network, driving their spiking activity. A pair of CCMs with opposite-signed inputs is employed for each quantity to allow for the encoding of both positive and negative magnitudes. We choose a Rectified Linear Unit (ReLU) transfer function for the CCMs to generate a positive current value when their input is positive while providing no output otherwise. In that way, the activity of the two CCM pairs is mutually exclusive since each quantity could be either positive or negative. The output currents of the CCMs are then passed to the sensory part of the network consisting of $PPC$ neurons. The $PPC$ neurons of the network responsible for sensing the control error and the rate of change in the control variable are then connected to the actuation component of the network, which updated the control variable to decrease the error.

\begin{figure}[!b]
\vspace{-5.2pt}
\centering
\includegraphics{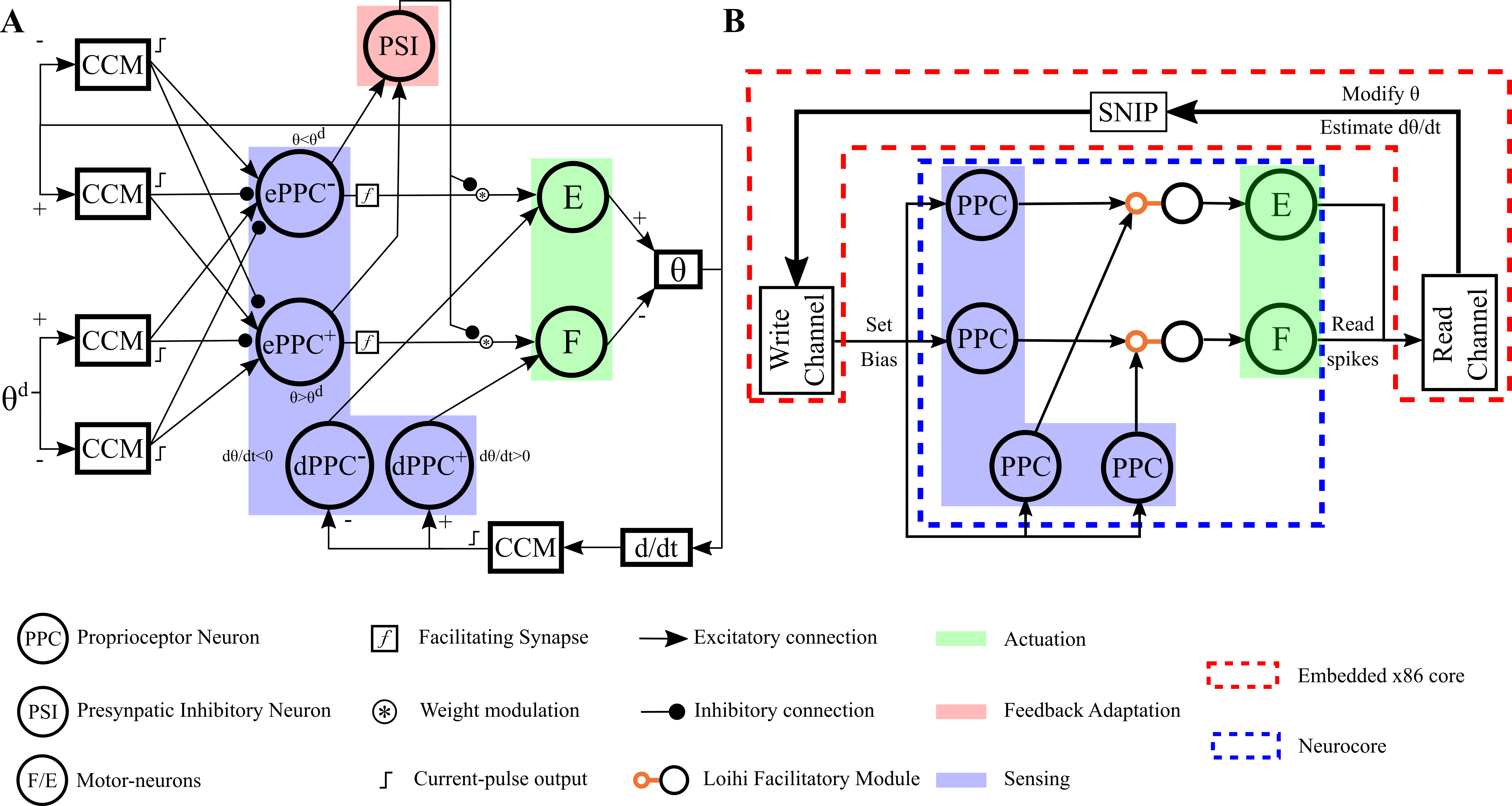}
\caption{Single-DOF SNN controller Architecture. CPU-based implementation (A) and deployment on Intel's Loihi (B). A pair of neurons ($ePPC$) compare the input target value ($\theta_d$) with the actual fed-back value of the controlled variable ($\theta$), and encode the difference in their activity. Then, they drive another pair of neurons (motor-neurons $F/E$) to in/de-crease $\theta$ and eliminate the difference. Short-term facilitation of the synaptic connections between the two pairs enforces gradual initiation of the elimination. A parallel pathway ($PSI$) also modifies the synaptic weights to adapt to a range of difference magnitudes. A third pair of neurons ($dPPC$) encodes the rate of change of $\theta$ to limit abrupt error corrections.}
\label{architecture}
\vspace{-5.2pt}
\end{figure}

\subsubsection{Sensing and Computing}
All neurons in the proposed SNN are simulated using the leaky-integrate-and-fire (LIF) model, governed by the following dynamics. First, we integrate the input spikes into synaptic current:

\begin{equation}
    u_{i}^{(t)} = u_{i}^{(t-1)}\cdot d_u + \sum_{j}^{} w_{ij}\cdot s_{j}^{(t-1)} 
\label{LIF1}
\end{equation}

Then, we integrate the synaptic current into the membrane voltage of the neuron, which then fires a spike when the voltage exceeded a threshold $V_{th}$, as follows:

\begin{equation}
\begin{split}
    v_{i}^{(t)} = v_{i}^{(t-1)}\cdot d_v + u_i^{(t)}, \text{if}\ v_{i}^{(t-1)} < V_{th} \\
    v_{i}^{(t)} = 0 \ \text{and}\ s_{i}^{t} = 1, \text{if}\ v_{i}^{(t-1)} \geq V_{th}
\end{split}
\label{LIF2}
\end{equation}

where $t$ is the time step, $u_i$ is the neuron's input current, $v_i$ is the neuron's voltage, $d_u$ and $d_v$ are current and voltage decay factors, $w_{ij}$ are the connection weights between the presynaptic neuron $j$ and the postsynaptic neuron $i$, and $s_j^{(t-1)}$ is a binary variable denoting the spikes of neuron $j$ at time step $t-1$.

To estimate the control error (difference between actual and desired value) at each timestep, we employ a pair of $PPC$ neurons ($ePPC$). These two neurons receive the current-encoded values of $\theta$ and $\theta^d$ as input and encode the sign and magnitude of the error in their spiking activity. To encode the sign, we connect the four CCMs to the pair of $ePPC$ in a way that enforces the mutually exclusive firing of the neurons. For example, when $0<\theta<\theta^d$, only the two middle CCMs produced non-zero output, with the output of the bottom one being larger in magnitude. Connecting the two CCMs with both $ePPC$ but with opposite signs ensures that $ePPC^-$ is excited, while the bottom one is shunted. In that way, $ePPC^-$ ($ePPC^+$) is always active (inactive) when $\theta<\theta^d$. Following the same logic, we choose the rest of the connections between the CCM and the $ePPC$. The magnitude of the error is reflected by the firing rate of the respective $ePPC$. The aggregate current from the CCM that stimulates the $ePPC$ was proportional to the error, inducing a proportional firing rate in the neurons. In the above example, $ePPC^-$ is stimulated by a positive current proportional to $\theta^d$ and a negative current proportional to $\theta$. Therefore, its firing rate is proportional to $\theta^d - \theta$. 

To improve the control smoothness, we utilize another pair of $PPC$ neurons ($dPPC$) that are sensitive to the rate of change of the control variable. These neurons introduce some push-pull dynamics: when $\theta$ diverges considerably from $\theta^d$, the controller attempts to quickly decrease the error, giving rise to a sizeable derivative value. This value is input to the respective $dPPC$ and activates it to counteract the error decrease and avoid overshooting. To incorporate the sensitivity of the $dPPC$ to the rate of change of the control variable, we first estimate the derivative of $\theta$ (measure its change over a time window and divide by the window length). Then, we encode its value into a current, which iss fed to the two $dPPC$ with opposite signs. In that way, these neurons fire in a mutually exclusive manner to encode positive and negative rates of change, with their firing rate encoding the magnitude of the derivative.

\subsubsection{Actuation}
The actuation component of the network comprises of two antagonistic motor neurons, an extensor (E) and a flexor (F), that increase and decrease the control variable, respectively. The E integrates the spikes of the top $ePPC$, whose activity indicates a negative error ($\theta < \theta^d$), and responds with spikes that increase $\theta$ towards the target $\theta^d$. For this, we translate each E spike to an incremental increase of $\theta$. Concurrently, the E integrates the spikes of the left $dPPC$, whose activity indicates a decrease in $\theta$ and responds with spikes to counteract that decrease. The connectivity and the function of the F neuron is analogous for the decrease of the control variable.

\subsubsection{Weight adaptation}
\label{sec:wgt_adapt}
To improve the motion smoothness without sacrificing control accuracy, we introduce two time-dependent, multiplicative factors $f^{(t)}$ and $g^{(t)}$ that adapt the static synaptic weights $W$ between the $ePPC$ and the motor neurons, as shown below:

\begin{equation}
w^{(t)} = W\cdot f^{(t)}\cdot g^{(t)}, 
\label{adaptation}
\end{equation}

First, we introduce a biologically plausible, activity-dependent facilitation of the synaptic weights~\cite{tsodyks1998neural} to maximize the smoothness of the motion. The efficacy $f_{ij}^{(t)}$ of the synaptic connection between the the $j^{th}$ $ePPC$ and the $i^{th}$ motor neuron increases with each presynaptic spike until saturating at a maximum value and decays otherwise with a factor $d_{fac}$, as shown in Eq. \ref{fac}. 

\begin{equation}
f_{ij}^{(t)} = f_{ij}^{(t-1)}\cdot d_{fac} + U_{fac}\cdot s_j^{(t-1)}, 
\label{fac}
\end{equation}
where $d_{fac}=1$ is the facilitation decay time constant (set to $500 ms^{-1}$), $U_{fac}$ is the facilitation factor increment (set to 0.005), and $s_j^{(t-1)}$ is the spike of the presynaptic neuron. Due to the low initial synaptic efficacies, abrupt increases in the activity of the $ePPC$ do not directly propagate to the motor neurons. The gradual increase of the synaptic efficacies results in a graded activation of the motor neurons, resulting in a smooth modification of the controlled amount. The parameter values for $d_{fac}$ and $U_{fac}$ are chosen to provide gradual movement initiation for the given arms, and can be modified over a range to increase/decrease this effect.

Second, we optimize the controller parameters according to the range of the desired movement to ensure accurate control for movements of significantly different magnitudes. Specifically, while small errors require strong synaptic connections from the $ePPC$ to activate the motor neurons and ensure accurate control, large errors require weaker connections to avoid overshoot. To address this requirement, we introduce the biologically plausible adaptation mechanism of presynaptic inhibition~\cite{fink2014presynaptic}, and scale the synaptic weights as a function of the error magnitude. Specifically, we utilize an additional neuron (PSI), whose activity reflects the activity of the $ePPC$ and, therefore, the magnitude of the error. The PSI spikes modulate the connection weights through a factor $g$ as described below: 

\begin{equation}
g^{(t)} = g_{max} - \Big[g^{(t-1)}\cdot d_{PSI} + U_{PSI}\cdot s_{PSI}^{(t-1)}\Big],
\label{PSI}
\end{equation}
where $g_{max}$ is the maximum value of the modulatory factor $g$ (set to 1), $d_{fac}=1$ is the presynaptic inhibition decay time constant (set to $50 ms^{-1}$), $U_{PSI}$ is the presynaptic inhibition increment (set to 0.005), and $s_{PSI}^{(t-1)}$ is the spike of the dedicated presynaptic inhibitory neuron. The parameter values for $d_{fac}$ and $U_{fac}$ are chosen to provide accurate control for joint movements in the maximum range allowed for the given arms, and can also be modified if this joint range were to be adapted to a different task. During the control, $g$ decreases by $U_{PSI}$ with each PSI spike $s_{PSI}^{(t-1)}$ and otherwise decays to its maximum value $g_{max}$ with a factor $d_{PSI}$. Using $g$ as a multiplicative factor in Eq. \ref{adaptation}, we emulate the divisive effect of presynaptic inhibition on the weights of biological synapses~\cite{fink2014presynaptic}.

\subsection{Neuromorphic Realization}
Due to the proposed algorithm's biological plausibility, the SNN presented can be easily and directly ported to any of the state-of-the-art neuromorphic platforms~\cite{davies2018loihi, furber2014spinnaker, moradi2017scalable, pehle2022brainscales} (Fig. \ref{architecture}, B). We deploy the SNN controller on Intel's Loihi neuromorphic research chip~\cite{davies2018loihi}, which we select because the chip's hardware architecture includes customizable, tree-like neuron models that allow for easy deployment of the proposed mechanisms (e.g. facilitation) We implement $PPC$ and actuation neurons as single Loihi compartments governed by the Leaky Integrate and Fire (LIF) dynamics (Eq. \ref{LIF1}-\ref{LIF2}) that Loihi inherently supports.

One limitation of the Loihi hardware is that it does not directly support facilitatory synapses. To overcome this limitation, we develop a facilitatory module. We design this module as a two-compartment neuron that follows the imbalanced tree structure of multi-compartment neurons on Loihi. The leaf (Fig. \ref{architecture}, B, small orange circle) is a non-spiking compartment that receives and integrates the module's input. The root (large black circle) is a spiking compartment that integrates the leaf compartment's voltage. This cascaded integration of the input results in a gradual increase in the spiking activity of the root compartment, which induces gradual activation of the motor neurons, approximating the effects of the facilitatory synapses in the off-chip controller.

While this design introduces more Loihi compartments and connections than the number of neurons and synapses in the CPU-based implementation, the single-joint controller only requires 10 compartments and 8 connections (60 compartments and 66 connections including the connections providing joint coordination for the control of the real-world Jaco arm). Because of its network size, the hardware implementation only utilizes one out of the 128 available Loihi neurocores, while similar methods~\cite{stagsted2020event} have to distribute their networks over 38 cores. Therefore, the biological plausibility of the architecture minimizes the resources requirements of the design.   

\begin{figure}[!t]
\vspace{-5.2pt}
\centering
\includegraphics[scale=1.9]{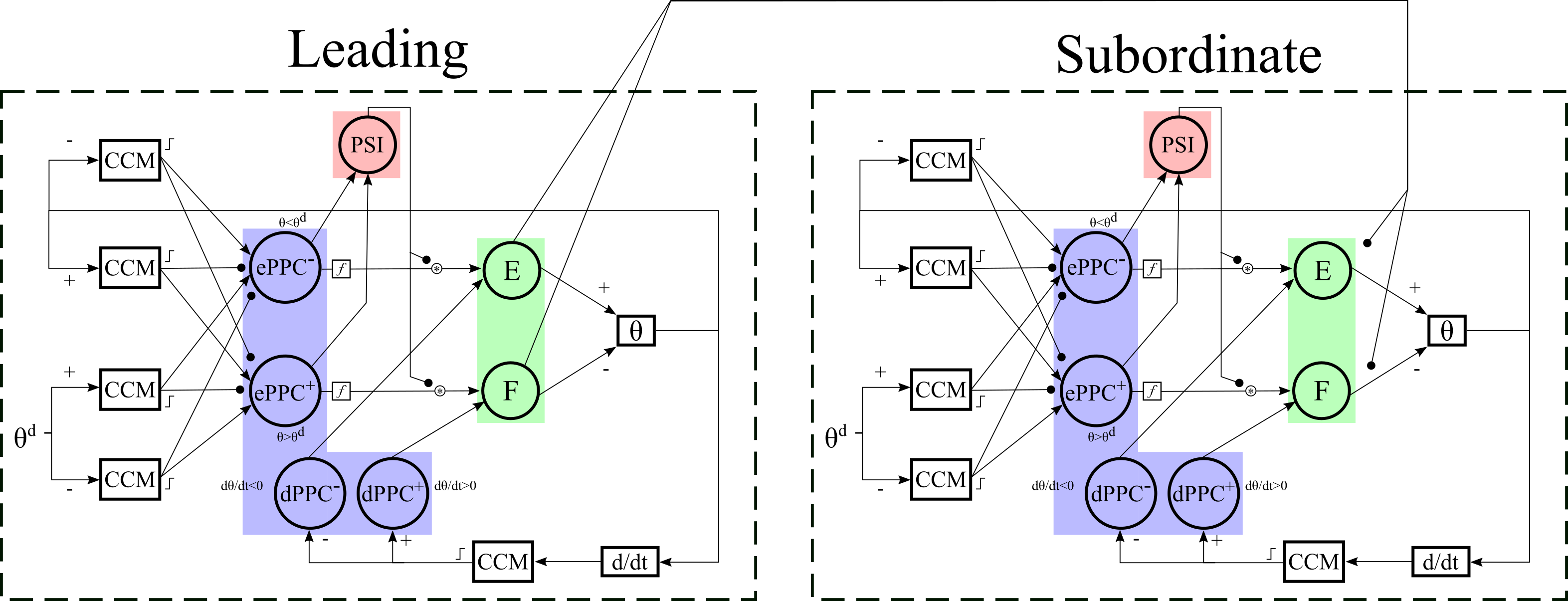}
\caption{Coordination of independent joint controllers. The motor activation in the leading joints inhibits the activation of the subordinate joints. In this way, the motion initiation in the subordinate joints is temporally delayed, allowing those joints to exploit the inertia of the leading joint's motion, as observed in human subjects~\cite{dounskaia2005internal}.}
\label{coordination}
\vspace{-5.2pt}
\end{figure}

\subsection{Multi-DOF Control}
To scale up the single-DOF control approach and allow for the control of real-world robotic arms with multiple DOF, we use the SNN as a control block and independently apply it to each DOF of the arm’s joints. We draw inspiration from the experimentally observed joint coordination in human arms~\cite{buchanan2004learning, dounskaia2005internal} and apply these principles in the proposed arm controllers to improve the control behavior (Fig. \ref{coordination}). Specifically, we incorporate the leading joint hypothesis~\cite{dounskaia2005internal}, which postulates that in human arms, the joints that are closer to the body lead the movement, and the ones further down the limb (subordinate) follow. In that way, the subordinate joints exploit the leading joint's inertia. To account for this, we stimulate the motor neurons of the subordinate joint with the spikes of the leading joint motor neurons. For the Reacher-v2, we define the first joint (shoulder) as leading and the second (elbow) as subordinate. For the Jaco arm, we define the first and second joints as leading and the remaining four as subordinate.

\begin{figure}[!b]
\vspace{-5.2pt}
\centering
\includegraphics[scale=1.5]{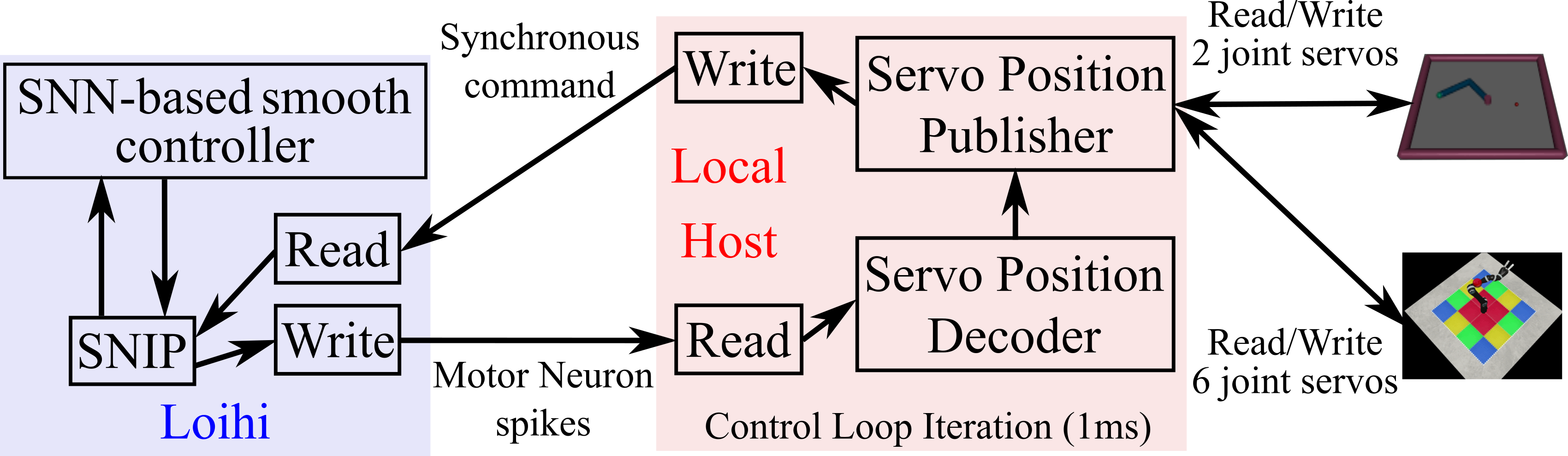}
\caption{Loihi-robot interaction control loop. Loihi communicates with the Local Host through a Read and a Write channel controlled by the on-chip x86 core (SNIP). The neuron spikes are decoded and used to update the simulated servos. The servo states are read out and fed back to Loihi.}
\label{control_loop}
\vspace{-5.2pt}
\end{figure}

\subsection{Controller-Robot Interaction}
To allow the SNN to control the robot arms, we translate each spike of the motor neurons to an incremental change (positive for E and negative for F) in the angle of the respective joint. For the Reacher-v2, we first ran a calibration simulation to determine the nominal torques that are required to change each joint angle by the desired increment. Then, we translate each $E/F$ spike during the arm control to a positive/negative torque increment of that magnitude. For the Jaco2 arm, we use the SNN to control the angle of each joint in the configuration space. Here, we predefine an angle increment and translate each spike of $E/F$ to an increase/decrease of the respective angle by that increment. Then, we translate the instructed angles to torques using an analytic operational space controller~\cite{khatib1987unified}. 

The interaction between Loihi and the simulation environments is realized using the framework proposed in~\cite{tang2020reinforcement} to allow for real-time control of the robot arms (Fig. \ref{control_loop}). For this, we use a Read channel controlled by the low-frequency x86 Loihi cores (SNIP) to read out the spiking activity of the motor neurons and translate it to joint movements. In real-time, estimates of $\theta$ and $d\theta/dt$ are fed as inputs to the corresponding $PPC$ neurons through a Write channel, which sets the biases of these $PPC$ neurons. 

\begin{figure}[!b]
\vspace{-5.2pt}
\centering
\includegraphics{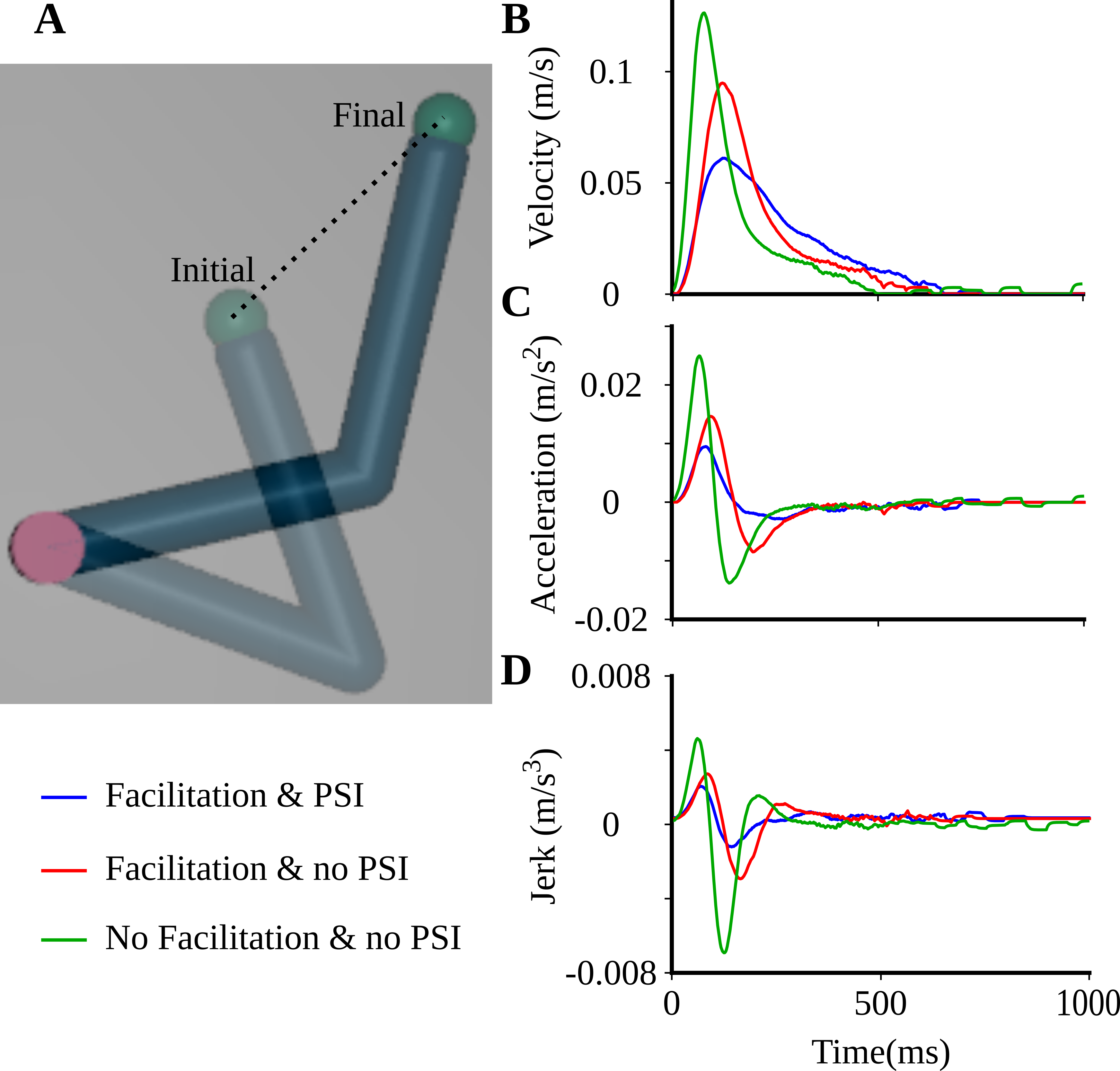}
\caption{Contribution of each network component to the improvement of the control smoothness. The SNN controller drove the 2-link arm from an initial to a final configuration (A). Without facilitation and/or presynaptic inhibition (green and red) successfully drives the discrete movement, but gives rise to fast and abrupt end-effector movements (B, C). The complete SNN controller (blue) maximizes (minimizes) the control smoothness (jerk) (D).}
\label{R1}
\vspace{-5.2pt}
\end{figure}

%% file: tex_input/results.tex
\subsection{Demonstrating smooth control and identifying its origins}
Providing smooth control is one of the main goals of our design. As a proof of concept, we use our SNN controller to drive a discrete movement of the end effector of the Reacher-v2 arm (Fig. \ref{R1}, A). To quantify the smoothness of our control approach, we compute a well-established smoothness metric~\cite{hogan1984organizing} the motion jerk (the second derivative of the motion velocity Fig. \ref{R1}). We show the recorded trajectory and the evaluation metrics (joint velocity, acceleration, and jerk) after a single simulation because the absence of stochastic components in the proposed controller does not introduce any intra-session variability. Mimicking the behavior of its biological counterpart~\cite{fink2014presynaptic}, the SNN controller gives rise to bell-shaped velocity profiles (Fig. \ref{R1}, B). The gradual acceleration and timely deceleration of the end-effector (Fig. \ref{R1}, C) confined the magnitude of the motion jerk (Fig. \ref{R1}, D), suggesting that the proposed SNN indeed provides smooth control of the robotic arm.

To determine the contribution of each component of the SNN to the desired control behavior, we sequentially ablate the mechanisms and evaluate the control dynamics. First, we deactivate the adaptation mechanism of PSI and use the modified SNN controller to drive the exact same discrete movement. While the initial phase of the movement is unaffected in the absence of PSI, the peak end-effector velocity is not adequately confined (Fig. \ref{R1}, B, red vs. blue curves). The deceleration from this higher peak velocity is more abrupt (Fig. \ref{R1}, C), leading to an increase in the motion jerk (both in terms of absolute magnitude and integral). Next, we deactivate both the PSI and the synaptic facilitation mechanisms, and the remaining SNN drives the same discrete movement. In the absence of the two mechanisms, both the increase and the decrease in the end-effector speed are abrupt, and the peak velocity is even higher (Fig. \ref{R1}, B, green vs. blue curve). The consequent larger values of acceleration  (Fig. \ref{R1}, C) result in significantly increased motion jerk (Fig. \ref{R1}, D).

\begin{figure}[!b]
\vspace{-15.2pt}
\centering
\includegraphics[scale=1.2]{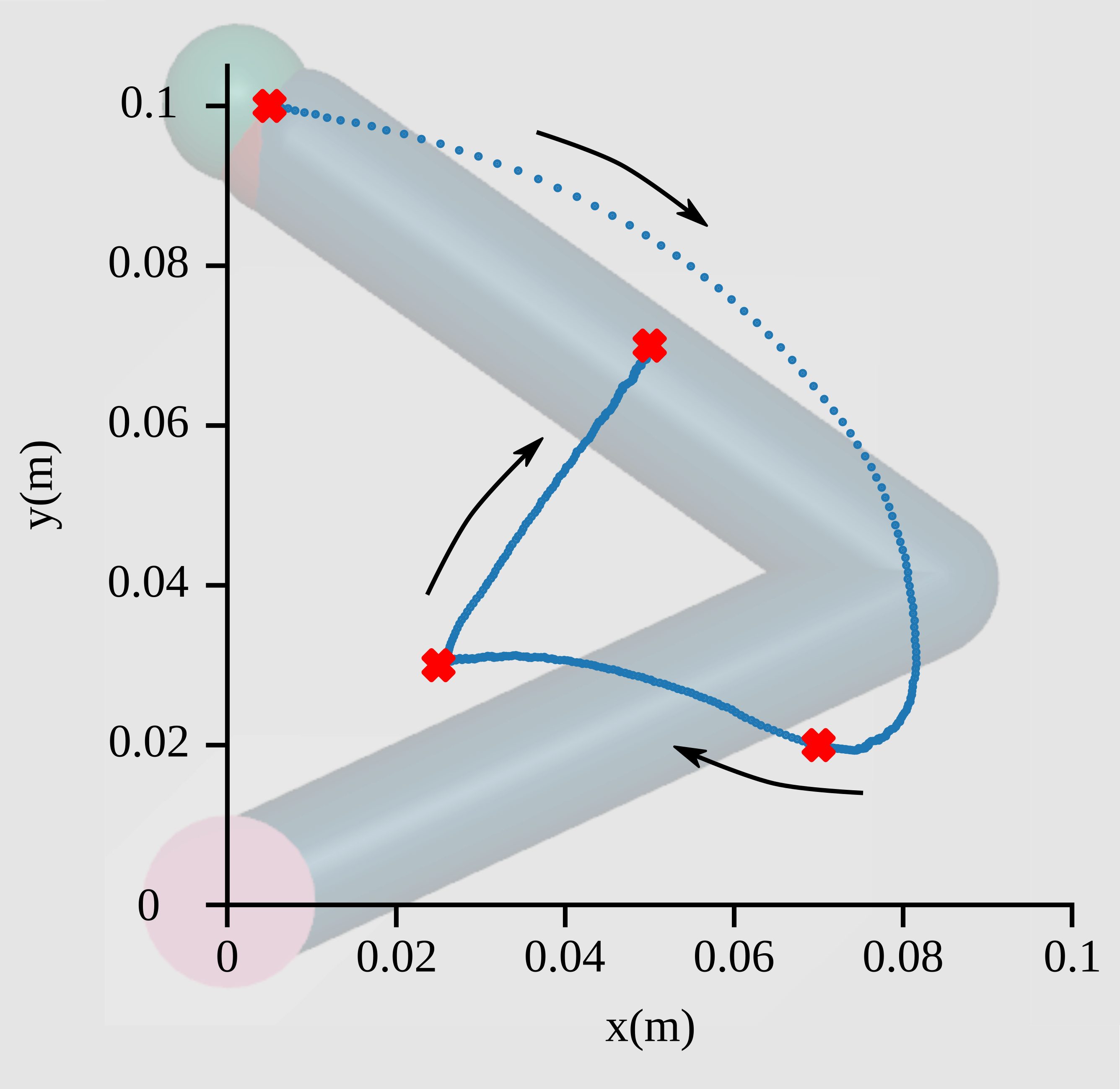}
\caption{End-effector trajectory for the 2-link arm when driven by the SNN controller to perform discrete movements. The curvature of the trajectories is in agreement with experimental evidence in human arms (see~\cite{konczak1997development}, Fig.2)}
\label{R4}
\vspace{-5.2pt}
\end{figure}

\subsection{Biological relevance of the motion trajectories}
The biological inspiration of the SNN architecture made us wonder whether the resultant arm movements are similar to those observed in human arms. To examine this, we apply the controller to the Reacher-v2 arm to drive a sequence of discrete movements and record the trajectories of its end-effector (Fig. \ref{R4}). At predefined intervals and regardless of the completion of the previous movements, a new, random goal position is defined for the end-effector. This workspace position is provided to an Inverse Kinematics (IK) solver that calculates the joint configuration that would drive the robot arm to the desired position. Then, the resulting angles are fed to the SNN controllers of the respective joints to drive their movements. As expected, the SNN controller provides smooth control with a gradual initial increase and timely decrease in the end-effector speed (as shown by the increased point density around the targets). Interestingly, the shape of the trajectories resembles the sigmoidal trajectories with curvatures in the early and late stages of the movement interrupted by straight intermediate intervals as observed in human subjects during adulthood and shown in~\cite{konczak1997development}. Considering that the proposed controller does not include motion planning beyond the calculation of the initial and final joint configurations, this natural emergence of slightly curved trajectories suggests that this human-like behavior results from the biologically plausible SNN architecture.

\begin{figure}[!t]
\vspace{+5.2pt}
\centering
\includegraphics[scale=1]{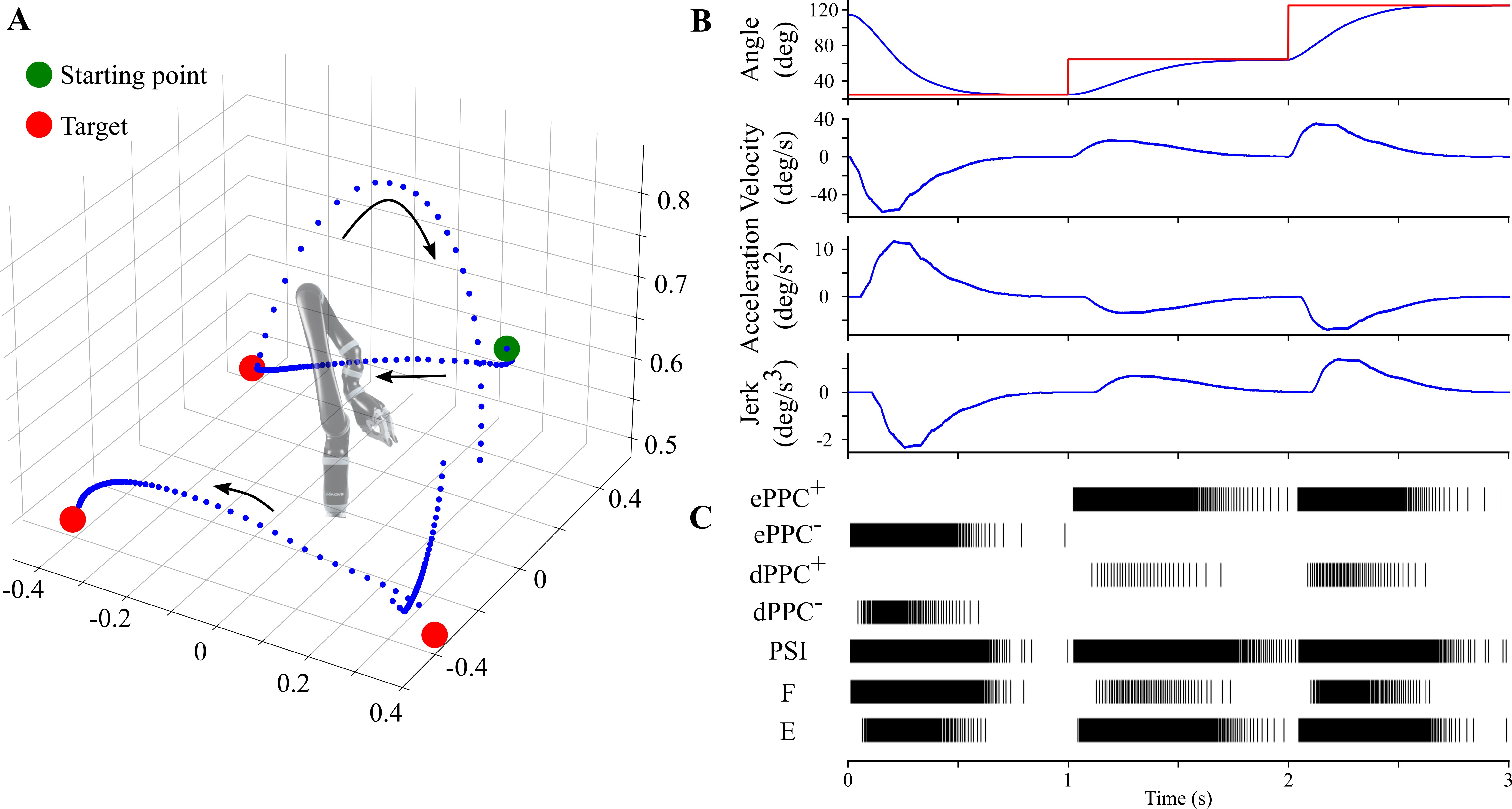}
\caption{SNN controller drives a Jaco arm. The end-effector trajectory of the of the Jaco arm while performing discrete movements in the configuration space exhibits increased density of points around the targets, demonstrating the smooth acceleration and deceleration in the early and late stages of the movements (A). The control dynamics of an exemplary arm joint (B) as a result of the neuronal spiking activity (C) of the SNN that controls the joint.}
\label{R2}
\vspace{-15.2pt}
\end{figure}

\subsection{Application to multi-DOF robot arms}
To demonstrate the applicability of the approach to real-world robotic arms, we scale up the proposed SNN to control a Jaco arm with 6 DOF. We use one controller block for each joint and introduce joint coordination as previously described. Here, the first joint of the arm acts as leading for the second, and both of them act as leading for the other four. In that way, we exploit the inertia of the first joint in the x-y plane and the inertia of the second joint along the z-axis. We utilize the SNN controller to drive the discrete movement of the arm’s end-effector between three points in the workspace and record its trajectory (Fig. \ref{R2}, A). As in the Reacher-v2 experiment, a new, random goal position for the end-effector is generated at predefined time instances and regardless of the completion of previous movements. Then, the IK solver for the Jaco arm transforms the workspace positions to joint configurations, and the target joints are fed to the respective controllers to drive the arm to the desired position. As discussed above, we only show the results of a single simulation since there was no variability across simulations. As in the case of the Reacher-v2 arm, the end-effector trajectory is smooth, with gradual deceleration when approaching the targets. This behavior of the end-effector is a result of the smooth control of each joint separately. As an example, we show the dynamics of one arm joint controlled by the proposed SNN (Fig. \ref{R2}, B). The joint moves accurately and smoothly to the desired angles without overshoot (Fig. \ref{R2}, B, Angle) while preserving the bell-shaped angular speed profiles (Fig. \ref{R2}, B, Velocity middle panel and continuous accelerations (Fig. \ref{R2}, B, Acceleration) that keep the motion jerk to minimal levels (Fig. \ref{R2}, B, Jerk). The spiking activity of the neurons in the respective SNN controller block explains the control dynamics of the single joint (Fig. \ref{R2}, C). Specifically, for each step movement, one of the $ePPCs$ is activated by the position error and drives its respective motor neuron ($E/F$) to decrease it. As the joint starts moving towards the target, the respective $dPPC$ is activated due to the change in the angular speed and, in turn, stimulates the antagonist motor neuron ($F/E$). This counteraction results in a timely joint deceleration when approaching the target, contributing to the smoothness of the control. Lastly, the PSI neuron contributes to the fine-tuning of the movement: This neuron is highly active in the early stages of the movement when the positional error is high. In that way, it downscales the synaptic weight between the $ePPCs$ and the motor neurons to avoid the overshoot induced by strong $E/F$ activation. As the error decreases towards the end of the movement, the activity of PSI fades, allowing for the gradual increase of the synaptic weight to its maximal value. The larger weight is then sufficient to accurately correct the small positional errors without inducing overshoots.

\begin{figure}[!t]
\vspace{-5.2pt}
\centering
\includegraphics[scale=1.2]{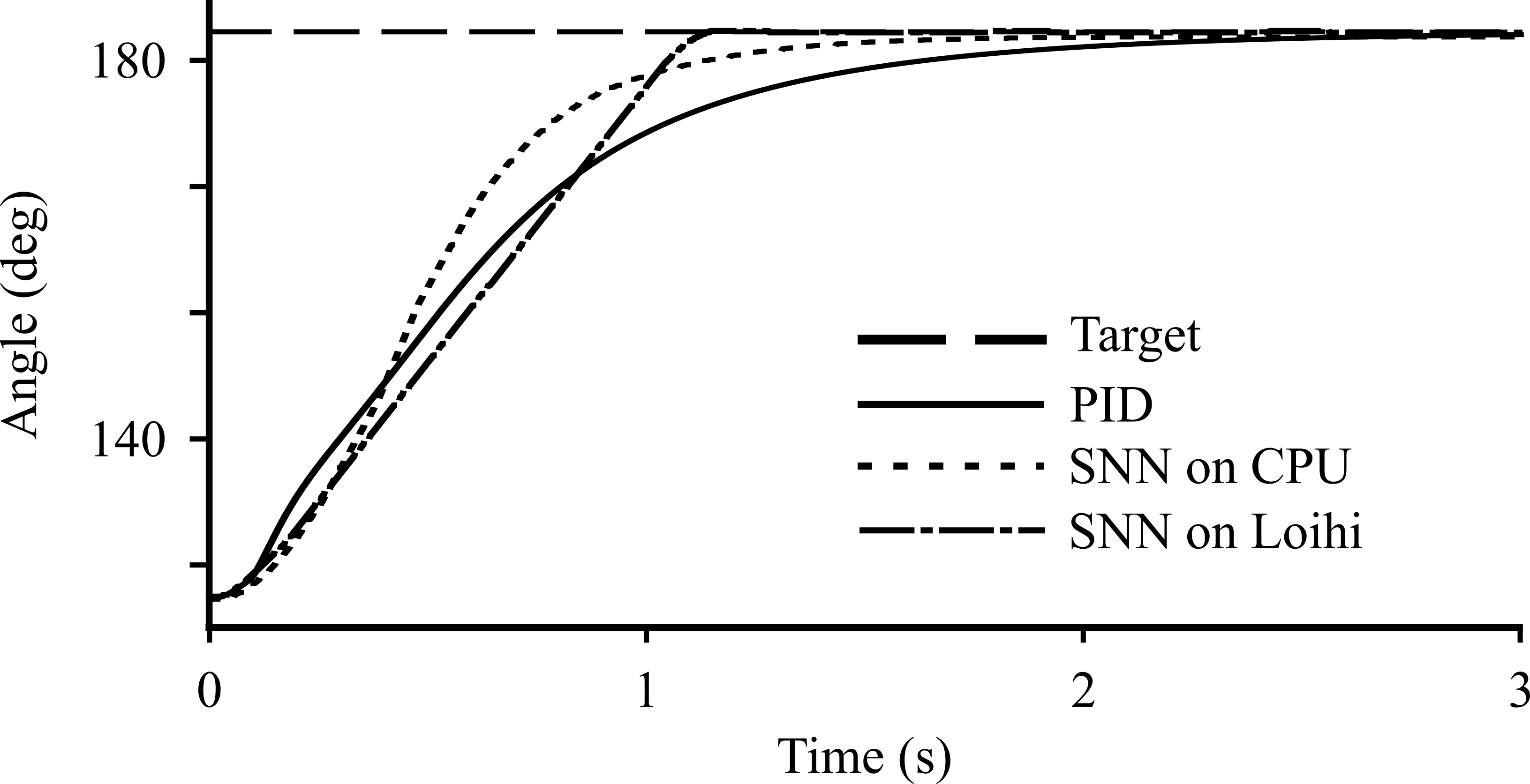}
\caption{Performance comparison between the SNN and a PID controller. Both controllers give rise to similar step responses when controlling a single joint of the Jaco arm. Quantitative comparison of the different controller metrics are shown in Table I.}
\label{R3}
\vspace{-15.2pt}
\end{figure}

\subsection{Comparison with conventional PID control}
After demonstrating that the proposed SNN controller is indeed applicable to real-world robotic arms with multiple DOF, we evaluate the approach by comparing it with a state-of-the-art PID controller that is widely used in similar applications. Specifically, we use as the baseline the operational space controller proposed in~\cite{khatib1987unified}, which is optimized for the control of the Jaco arm in the Applied Brain Research control library \footnote{The Applied brain research control library can be found at: \url{https://github.com/abr/abr_control}. Copyright (c) 2017-2022 Applied Brain Research. Used with permission.}. The set of tuned PID parameters are $K_p = 20$ for the proportional component, $K_v = 30$ for the derivative component, and $K_i = 0.001$ for the integral component of the position, while $K_o = 180$ is set for the proportional control of the orientation. We then tune the SNN controller's parameters to approximate the control profiles (convergence time, accuracy).

We compare the step response of the two controllers when required to drive a single robot joint from an initial to a desired configuration (Fig. \ref{R3}). As shown by the control performance of the first leading joint (Fig. \ref{R3}), the SNN controller implemented on either a CPU or Loihi achieves comparable performance to the PID. We quantify this comparison by measuring the overshoot, the rise time, and the settling time of the controllers (Table \ref{metrics})~\cite{levine2018control}. The SNN controllers exhibit a similar rise and settling time with the PID. While the SNN controllers are slightly faster than the PID, we clarify that the goal is not to outperform the conventional control method but to achieve comparable results with a biologically plausible and, therefore, neuromorphic-compatible approach. When implemented on Loihi, the SNN controller exhibits a small (1.7\%) overshoot which is absent in the CPU implementation. This is expected due to the absence of presynaptic inhibition in the Loihi deployment, which further highlights the contribution of this adaptation mechanism to the control.

\begin{figure}[!b]
\vspace{+5.2pt}
\centering
\includegraphics[scale=0.9]{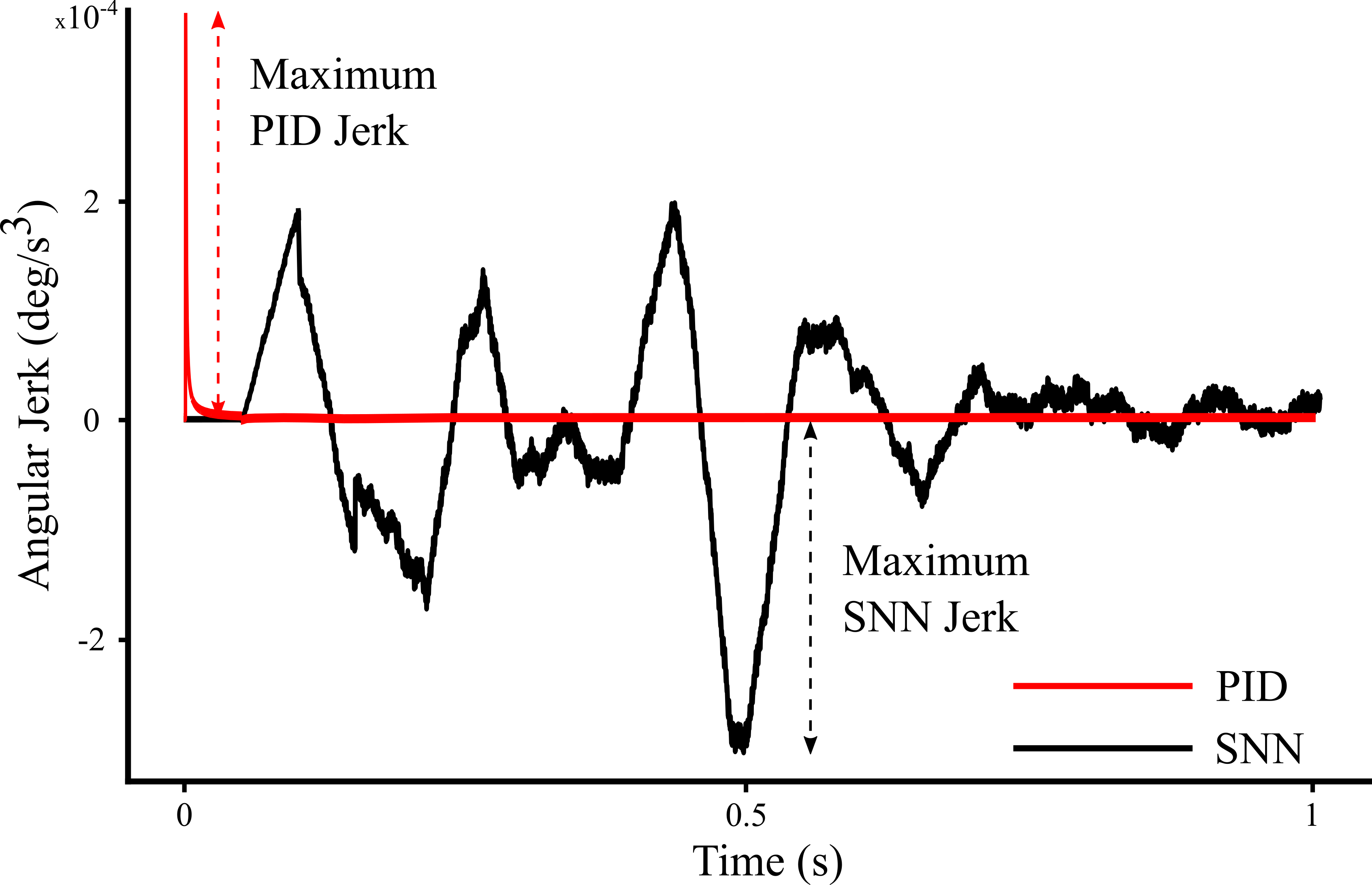}
\caption{Comparison of the end-effector's jerk when driven by the SNN (black) and a PID (red) controller. The maximal jerk was lower for the SNN controller resulting in more smooth motion.}
\label{R5}
\vspace{-15.2pt}
\end{figure}

With motion smoothness being the main goal for the proposed controller design, we further examine this aspect of its performance and compare it against the state-of-the-art PID controller. Both controllers drive the same discrete movement of the Jaco arm (Section IV.D) and we evaluate the motion smoothness (Fig. \ref{R5}). To do so, we use the well-established jerk metric (derivative of the acceleration)~\cite{kyriakopoulos1988minimum}, which is minimal in smooth human arm movements~\cite{flash1985coordination}. When using the PID controller, an initial spike dominates the motion jerk, which quickly decays to zero subsequently. On the other hand, the SNN controller gives rise to a gradual increase in the motion jerk that persists longer, decreasing the maximum jerk by 19\%. With jerk quantifying how abrupt the movement is, high instantaneous values deteriorate the movement smoothness more than smaller and persistent ones. Therefore, the proposed method indeed improves motion smoothness, approximating the minimum jerk profiles in human arms.         
\begin{table}[t!]
\caption{Controller Metrics Comparison.}
\begin{indented}
\item[]
\begin{tabular}{|c||c|c|c|}
\hline
                                                      & \begin{tabular}[c]{@{}c@{}}Overshoot\\ ($\theta_{max} - \theta_{des}$ \%)\end{tabular} & \begin{tabular}[c]{@{}c@{}}Rise time\\ (10\% - 90\% $\theta_{des}$)\end{tabular} & \begin{tabular}[c]{@{}c@{}}Settling Time\\ ($\textless{}20\% \theta_{des}$)\end{tabular} \\ \hline \hline
PID                                                   & 0                                                                                & 1136.9 ms                                                                     & 945.5 ms                                                                             \\ \hline
\begin{tabular}[c]{@{}c@{}}SNN (CPU)\end{tabular}   & 0                                                                                & 724.3 ms                                                                      & 723.8 ms                                                                             \\ \hline
\begin{tabular}[c]{@{}c@{}}SNN (Loihi)\end{tabular} & 1.7                                                                              & 804 ms                                                                        & 881 ms                                                                               \\ \hline
\end{tabular}
\end{indented}
\label{metrics}
\end{table}

%% file: tex_input/discussion.tex
In this paper, we presented an SNN controller which draws inspiration from biological findings on how primates perform smooth reaching movements~\cite{fink2014presynaptic}. We evaluated our controller on two robotic arms: i) the simple but human-like two-link arm Reacher-v2, and ii) the real-world 6-DOF Jaco arm. Our results show that the controller not only drove the arm to the target accurately but also provided smooth control, thus making it applicable for a wide variety of robotic applications, such as prosthetics~\cite{hochberg2006neuronal}, assistive arms~\cite{naotunna2015meal}, or industrial arms~\cite{xiao2012smooth}, that may require such control. 

Adding to the mounting evidence that demonstrates the role of neuroscience in advancing robotics~\cite{sandamirskaya2022neuromorphic, dupeyroux2021toolbox, manoonpong2013neural, polykretis2020astrocyte, ijspeert2008central}, we show here how the emulation of neuronal subnetworks with specialized functions can provide similar behavior on neuromorphic-controlled robots. Several recent studies have attempted to harness the advantages of neuromorphic platforms for low-level control~\cite{christensenevent, zhao2020closed,zaidel2021neuromorphic}. Their performance approximates that of the conventional microcontrollers, showing that the non-Von Neumann architecture of the hardware is not a compounding factor. However, the deployment of conventional control methods on neuromorphic hardware disregards the underlying neuroscience principles. As a result, such methods only partially exploit the low complexity of the minimal brain networks. Here, we propose an SNN controller that emulates (i) the experimentally identified GABA-ergic pathways that regulate the gain of sensory feedback afferents, and (ii) the short-term synaptic plasticity that enforces gradual motion initiation. The combination of these two mechanisms in our SNN architecture results in smooth joint control that resembles the dynamics of reaching movements in primates~\cite{seki2003sensory} and rodents~\cite{fink2014presynaptic}. This biological plausibility not only resulted in a minimal network in terms of size and complexity but also made it a perfect fit for deployment on neuromorphic hardware. Functionally, the controller inherited the behavior of its biological counterpart, as control smoothness emerged naturally through the gradual increase and timely and steady decline of the controlled variable.

While the biologically inspired design of our SNN controller resulted in the above mentioned architectural and functional advantages, it also gave rise to some limitations. First, our controller, similarly to other SNN solutions, lacks the concrete mathematical description of a PID controller. Drawing inspiration from biological networks controlling reaching movements, our SNN design is interpretable at the behavioral level. However, a thorough analysis of the control behavior is less straightforward when compared to conventional control methods. Second, our SNN controller introduced more tunable parameters than the three PID coefficients. Although the behaviorally interpretable nature of our design simplifies the tuning of the network hyperparameters, their increased number introduces additional complexity. Lastly, the binary nature of our spiking controller might introduce limited flexibility when compared with a continuous conventional controller. However, our method could be expanded with and benefit from previously proposed SNN controllers~\cite{jimenez2012neuro} that attempt to overcome this limitation.

Overall, this work supports ongoing efforts to develop real-time and robust neuromorphic solutions for robot control exhibiting desired behaviors~\cite{ehrlich2022adaptive}. The proposed methods could be combined with emerging solutions for the inverse-kinematics calculation of high-dimensional robotic arms~\cite{zhao2022learning, volinski2022data} to provide end-to-end neuromorphic control. Unlike other methods, here we aimed to derive computational principles from the knowledge of how the brain attains optimal motor behavior and introduce them to the design of neuromorphic robotic controllers, adding to the growing pool of non-cognitive neuromorphic applications~\cite{aimone2022review}. In the light of our results, which combine low computational complexity and interpretability with a performance comparable to the state-of-the-art, the direction of developing brain-inspired solutions to robotic problems on neuromorphic hardware~\cite{schuman2022opportunities} is worth exploring further.

%% file: main.bbl
\begin{thebibliography}{10}

\bibitem{aimone2022review}
J.~Aimone, P.~Date, G.~Fonseca-Guerra, K.~Hamilton, K.~Henke, B.~Kay,
  G.~Kenyon, S.~Kulkarni, S.~Mniszewski, M.~Parsa, et~al.
\newblock A review of non-cognitive applications for neuromorphic computing.
\newblock {\em Neuromorphic Computing and Engineering}, 2022.

\bibitem{ajoudani2018progress}
A.~Ajoudani, A.~M. Zanchettin, S.~Ivaldi, A.~Albu-Sch{\"a}ffer, K.~Kosuge, and
  O.~Khatib.
\newblock Progress and prospects of the human--robot collaboration.
\newblock {\em Autonomous Robots}, 42:957--975, 2018.

\bibitem{arber2012motor}
S.~Arber.
\newblock Motor circuits in action: specification, connectivity, and function.
\newblock {\em Neuron}, 74(6):975--989, 2012.

\bibitem{ata2007optimal}
A.~A. Ata.
\newblock Optimal trajectory planning of manipulators: a review.
\newblock {\em Journal of Engineering Science and technology}, 2(1):32--54,
  2007.

\bibitem{baker2007oscillatory}
S.~N. Baker.
\newblock Oscillatory interactions between sensorimotor cortex and the
  periphery.
\newblock {\em Current opinion in neurobiology}, 17(6):649--655, 2007.

\bibitem{balasubramanian2015analysis}
S.~Balasubramanian, A.~Melendez-Calderon, A.~Roby-Brami, and E.~Burdet.
\newblock On the analysis of movement smoothness.
\newblock {\em Journal of neuroengineering and rehabilitation}, 12(1):1--11,
  2015.

\bibitem{brogaardh2007present}
T.~Brog{\aa}rdh.
\newblock Present and future robot control development—an industrial
  perspective.
\newblock {\em Annual Reviews in Control}, 31(1):69--79, 2007.

\bibitem{buchanan2004learning}
J.~J. Buchanan.
\newblock Learning a single limb multijoint coordination pattern: the impact of
  a mechanical constraint on the coordination dynamics of learning and
  transfer.
\newblock {\em Experimental Brain Research}, 156(1):39--54, 2004.

\bibitem{calimera2013human}
A.~Calimera, E.~Macii, and M.~Poncino.
\newblock The human brain project and neuromorphic computing.
\newblock {\em Functional neurology}, 28(3):191, 2013.

\bibitem{capaday1987difference}
C.~Capaday and R.~Stein.
\newblock Difference in the amplitude of the human soleus h reflex during
  walking and running.
\newblock {\em The Journal of physiology}, 392(1):513--522, 1987.

\bibitem{capaday1987method}
C.~Capaday and R.~Stein.
\newblock A method for simulating the reflex output of a motoneuron pool.
\newblock {\em Journal of neuroscience methods}, 21(2-4):91--104, 1987.

\bibitem{christensenevent}
A.~L. Christensen and Y.~Sandamirskaya.
\newblock Event-based pid controller fully realized in neuromorphic hardware: a
  one dof study.

\bibitem{constantinescu2000smooth}
D.~Constantinescu and E.~A. Croft.
\newblock Smooth and time-optimal trajectory planning for industrial
  manipulators along specified paths.
\newblock {\em Journal of robotic systems}, 17(5):233--249, 2000.

\bibitem{crespi2005amphibot}
A.~Crespi, A.~Badertscher, A.~Guignard, and A.~J. Ijspeert.
\newblock Amphibot i: an amphibious snake-like robot.
\newblock {\em Robotics and Autonomous Systems}, 50(4):163--175, 2005.

\bibitem{davies2018loihi}
M.~Davies, N.~Srinivasa, T.-H. Lin, G.~Chinya, Y.~Cao, S.~H. Choday, G.~Dimou,
  P.~Joshi, N.~Imam, S.~Jain, et~al.
\newblock Loihi: A neuromorphic manycore processor with on-chip learning.
\newblock {\em IEEE Micro}, 38(1):82--99, 2018.

\bibitem{dewolf2023neuromorphic}
T.~DeWolf, K.~Patel, P.~Jaworski, R.~Leontie, J.~Hays, and C.~Eliasmith.
\newblock Neuromorphic control of a simulated 7-dof arm using loihi.
\newblock {\em Neuromorphic Computing and Engineering}, 2023.

\bibitem{dounskaia2005internal}
N.~Dounskaia.
\newblock The internal model and the leading joint hypothesis: implications for
  control of multi-joint movements.
\newblock {\em Experimental Brain Research}, 166(1):1--16, 2005.

\bibitem{dupeyroux2021toolbox}
J.~Dupeyroux, S.~Stroobants, and G.~de~Croon.
\newblock A toolbox for neuromorphic sensing in robotics.
\newblock {\em arXiv preprint arXiv:2103.02751}, 2021.

\bibitem{ehrlich2022adaptive}
M.~Ehrlich, Y.~Zaidel, P.~L. Weiss, A.~M. Yekel, N.~Gefen, L.~Supic, and E.~E.
  Tsur.
\newblock Adaptive control of a wheelchair mounted robotic arm with
  neuromorphically integrated velocity readings and online-learning.
\newblock {\em Frontiers in Neuroscience}, 16:1007736--1007736, 2022.

\bibitem{fink2014presynaptic}
A.~J. Fink, K.~R. Croce, Z.~J. Huang, L.~Abbott, T.~M. Jessell, and E.~Azim.
\newblock Presynaptic inhibition of spinal sensory feedback ensures smooth
  movement.
\newblock {\em Nature}, 509(7498):43--48, 2014.

\bibitem{flash1985coordination}
T.~Flash and N.~Hogan.
\newblock The coordination of arm movements: an experimentally confirmed
  mathematical model.
\newblock {\em Journal of neuroscience}, 5(7):1688--1703, 1985.

\bibitem{furber2014spinnaker}
S.~B. Furber, F.~Galluppi, S.~Temple, and L.~A. Plana.
\newblock The spinnaker project.
\newblock {\em Proceedings of the IEEE}, 102(5):652--665, 2014.

\bibitem{gasparetto2015path}
A.~Gasparetto, P.~Boscariol, A.~Lanzutti, and R.~Vidoni.
\newblock Path planning and trajectory planning algorithms: A general overview.
\newblock {\em Motion and Operation Planning of Robotic Systems: Background and
  Practical Approaches}, pages 3--27, 2015.

\bibitem{glatz2019adaptive}
S.~Glatz, J.~Martel, R.~Kreiser, N.~Qiao, and Y.~Sandamirskaya.
\newblock Adaptive motor control and learning in a spiking neural network
  realised on a mixed-signal neuromorphic processor.
\newblock In {\em 2019 International Conference on Robotics and Automation
  (ICRA)}, pages 9631--9637. IEEE, 2019.

\bibitem{grillner1998intrinsic}
S.~Grillner, {\"O}.~Ekeberg, A.~El~Manira, A.~Lansner, D.~Parker, J.~Tegner,
  and P.~Wallen.
\newblock Intrinsic function of a neuronal network—a vertebrate central
  pattern generator.
\newblock {\em Brain Research Reviews}, 26(2-3):184--197, 1998.

\bibitem{hochberg2006neuronal}
L.~R. Hochberg, M.~D. Serruya, G.~M. Friehs, J.~A. Mukand, M.~Saleh, A.~H.
  Caplan, A.~Branner, D.~Chen, R.~D. Penn, and J.~P. Donoghue.
\newblock Neuronal ensemble control of prosthetic devices by a human with
  tetraplegia.
\newblock {\em Nature}, 442(7099):164--171, 2006.

\bibitem{hogan1984organizing}
N.~Hogan.
\newblock An organizing principle for a class of voluntary movements.
\newblock {\em Journal of neuroscience}, 4(11):2745--2754, 1984.

\bibitem{ijspeert2008central}
A.~J. Ijspeert.
\newblock Central pattern generators for locomotion control in animals and
  robots: a review.
\newblock {\em Neural networks}, 21(4):642--653, 2008.

\bibitem{ijspeert2007swimming}
A.~J. Ijspeert, A.~Crespi, D.~Ryczko, and J.-M. Cabelguen.
\newblock From swimming to walking with a salamander robot driven by a spinal
  cord model.
\newblock {\em science}, 315(5817):1416--1420, 2007.

\bibitem{imam2020rapid}
N.~Imam and T.~A. Cleland.
\newblock Rapid online learning and robust recall in a neuromorphic olfactory
  circuit.
\newblock {\em Nature Machine Intelligence}, 2(3):181--191, 2020.

\bibitem{jimenez2012neuro}
A.~Jimenez-Fernandez, G.~Jimenez-Moreno, A.~Linares-Barranco, M.~J.
  Dominguez-Morales, R.~Paz-Vicente, and A.~Civit-Balcells.
\newblock A neuro-inspired spike-based pid motor controller for multi-motor
  robots with low cost fpgas.
\newblock {\em Sensors}, 12(4):3831--3856, 2012.

\bibitem{khatib1987unified}
O.~Khatib.
\newblock A unified approach for motion and force control of robot
  manipulators: The operational space formulation.
\newblock {\em IEEE Journal on Robotics and Automation}, 3(1):43--53, 1987.

\bibitem{konczak1997development}
J.~Konczak and J.~Dichgans.
\newblock The development toward stereotypic arm kinematics during reaching in
  the first 3 years of life.
\newblock {\em Experimental brain research}, 117(2):346--354, 1997.

\bibitem{kreiser2020chip}
R.~Kreiser, A.~Renner, V.~R. Leite, B.~Serhan, C.~Bartolozzi, A.~Glover, and
  Y.~Sandamirskaya.
\newblock An on-chip spiking neural network for estimation of the head pose of
  the icub robot.
\newblock {\em Frontiers in Neuroscience}, 14, 2020.

\bibitem{kyriakopoulos1988minimum}
K.~J. Kyriakopoulos and G.~N. Saridis.
\newblock Minimum jerk path generation.
\newblock In {\em Proceedings. 1988 IEEE international conference on robotics
  and automation}, pages 364--369. IEEE, 1988.

\bibitem{levine2018control}
W.~S. Levine.
\newblock {\em The Control Systems Handbook: Control System Advanced Methods}.
\newblock CRC press, 2018.

\bibitem{liu2016rapid}
X.~Liu and Z.~Shen.
\newblock Rapid smooth entry trajectory planning for high lift/drag hypersonic
  glide vehicles.
\newblock {\em Journal of Optimization Theory and Applications}, 168:917--943,
  2016.

\bibitem{luo1985lq}
G.~Luo and G.~Saridis.
\newblock Lq design of pid controllers for robot arms.
\newblock {\em IEEE Journal on Robotics and Automation}, 1(3):152--159, 1985.

\bibitem{malki1997fuzzy}
H.~A. Malki, D.~Misir, D.~Feigenspan, and G.~Chen.
\newblock Fuzzy pid control of a flexible-joint robot arm with uncertainties
  from time-varying loads.
\newblock {\em IEEE Transactions on Control Systems Technology}, 5(3):371--378,
  1997.

\bibitem{manoonpong2013neural}
P.~Manoonpong, U.~Parlitz, and F.~W{\"o}rg{\"o}tter.
\newblock Neural control and adaptive neural forward models for insect-like,
  energy-efficient, and adaptable locomotion of walking machines.
\newblock {\em Frontiers in neural circuits}, 7:12, 2013.

\bibitem{michaelis2020robust}
C.~Michaelis, A.~B. Lehr, and C.~Tetzlaff.
\newblock Robust trajectory generation for robotic control on the neuromorphic
  research chip loihi.
\newblock {\em Frontiers in neurorobotics}, 14:589532, 2020.

\bibitem{milde2017obstacle}
M.~B. Milde, H.~Blum, A.~Dietm{\"u}ller, D.~Sumislawska, J.~Conradt,
  G.~Indiveri, and Y.~Sandamirskaya.
\newblock Obstacle avoidance and target acquisition for robot navigation using
  a mixed signal analog/digital neuromorphic processing system.
\newblock {\em Frontiers in neurorobotics}, 11:28, 2017.

\bibitem{moradi2017scalable}
S.~Moradi, N.~Qiao, F.~Stefanini, and G.~Indiveri.
\newblock A scalable multicore architecture with heterogeneous memory
  structures for dynamic neuromorphic asynchronous processors (dynaps).
\newblock {\em IEEE transactions on biomedical circuits and systems},
  12(1):106--122, 2017.

\bibitem{nadim2000role}
F.~Nadim and Y.~Manor.
\newblock The role of short-term synaptic dynamics in motor control.
\newblock {\em Current opinion in neurobiology}, 10(6):683--690, 2000.

\bibitem{naotunna2015meal}
I.~Naotunna, C.~J. Perera, C.~Sandaruwan, R.~Gopura, and T.~D. Lalitharatne.
\newblock Meal assistance robots: A review on current status, challenges and
  future directions.
\newblock In {\em 2015 IEEE/SICE International Symposium on System Integration
  (SII)}, pages 211--216. IEEE, 2015.

\bibitem{pehle2022brainscales}
C.~Pehle, S.~Billaudelle, B.~Cramer, J.~Kaiser, K.~Schreiber, Y.~Stradmann,
  J.~Weis, A.~Leibfried, E.~M{\"u}ller, and J.~Schemmel.
\newblock The brainscales-2 accelerated neuromorphic system with hybrid
  plasticity.
\newblock {\em Frontiers in Neuroscience}, 16, 2022.

\bibitem{polykretis2022spiking}
I.~Polykretis, G.~Tang, P.~Balachandar, and K.~P. Michmizos.
\newblock A spiking neural network mimics the oculomotor system to control a
  biomimetic robotic head without learning on a neuromorphic hardware.
\newblock {\em IEEE Transactions on Medical Robotics and Bionics}, pages 1--10,
  2022.

\bibitem{polykretis2020astrocyte}
I.~Polykretis, G.~Tang, and K.~P. Michmizos.
\newblock An astrocyte-modulated neuromorphic central pattern generator for
  hexapod robot locomotion on intel’s loihi.
\newblock In {\em International Conference on Neuromorphic Systems 2020}, pages
  1--9, 2020.

\bibitem{rivera2010extending}
J.~Rivera-Guillen, R.~Romero-Troncoso, A.~Osornio-Rios, A.~Garcia-Perez, and
  I.~Torres-Pacheco.
\newblock Extending tool-life through jerk-limited motion dynamics in machining
  processes: An experimental study.
\newblock 2010.

\bibitem{rossignol2006dynamic}
S.~Rossignol, R.~Dubuc, and J.-P. Gossard.
\newblock Dynamic sensorimotor interactions in locomotion.
\newblock {\em Physiological reviews}, 86(1):89--154, 2006.

\bibitem{sacrey2010development}
L.-A.~R. Sacrey and I.~Q. Whishaw.
\newblock Development of collection precedes targeted reaching: Resting shapes
  of the hands and digits in 1--6-month-old human infants.
\newblock {\em Behavioural Brain Research}, 214(1):125--129, 2010.

\bibitem{sandamirskaya2022neuromorphic}
Y.~Sandamirskaya, M.~Kaboli, J.~Conradt, and T.~Celikel.
\newblock Neuromorphic computing hardware and neural architectures for
  robotics.
\newblock {\em Science Robotics}, 7(67):eabl8419, 2022.

\bibitem{schlaghecken2002motor}
F.~Schlaghecken and M.~Eimer.
\newblock Motor activation with and without inhibition: Evidence for a
  threshold mechanism in motor control.
\newblock {\em Perception \& psychophysics}, 64(1):148--162, 2002.

\bibitem{schuman2022opportunities}
C.~D. Schuman, S.~R. Kulkarni, M.~Parsa, J.~P. Mitchell, B.~Kay, et~al.
\newblock Opportunities for neuromorphic computing algorithms and applications.
\newblock {\em Nature Computational Science}, 2(1):10--19, 2022.

\bibitem{seki2003sensory}
K.~Seki, S.~I. Perlmutter, and E.~E. Fetz.
\newblock Sensory input to primate spinal cord is presynaptically inhibited
  during voluntary movement.
\newblock {\em Nature neuroscience}, 6(12):1309--1316, 2003.

\bibitem{stagg2011role}
C.~J. Stagg, V.~Bachtiar, and H.~Johansen-Berg.
\newblock The role of gaba in human motor learning.
\newblock {\em Current biology}, 21(6):480--484, 2011.

\bibitem{stagsted2020towards}
R.~Stagsted, A.~Vitale, J.~Binz, L.~Bonde~Larsen, Y.~Sandamirskaya, et~al.
\newblock Towards neuromorphic control: A spiking neural network based pid
  controller for uav.
\newblock RSS, 2020.

\bibitem{stagsted2020event}
R.~K. Stagsted, A.~Vitale, A.~Renner, L.~B. Larsen, A.~L. Christensen, and
  Y.~Sandamirskaya.
\newblock Event-based pid controller fully realized in neuromorphic hardware: a
  one dof study.
\newblock In {\em 2020 IEEE/RSJ International Conference on Intelligent Robots
  and Systems (IROS)}, pages 10939--10944. IEEE, 2020.

\bibitem{strohmer2020flexible}
B.~Strohmer, P.~Manoonpong, and L.~B. Larsen.
\newblock Flexible spiking cpgs for online manipulation during hexapod walking.
\newblock {\em Frontiers in neurorobotics}, 14:41, 2020.

\bibitem{tan2011stable}
J.~Tan, K.~Liu, and G.~Turk.
\newblock Stable proportional-derivative controllers.
\newblock {\em IEEE Computer Graphics and Applications}, 31(4):34--44, 2011.

\bibitem{tang2020reinforcement}
G.~Tang, N.~Kumar, and K.~P. Michmizos.
\newblock Reinforcement co-learning of deep and spiking neural networks for
  energy-efficient mapless navigation with neuromorphic hardware.
\newblock {\em arXiv preprint arXiv:2003.01157}, 2020.

\bibitem{tang2019spiking}
G.~Tang, A.~Shah, and K.~P. Michmizos.
\newblock Spiking neural network on neuromorphic hardware for energy-efficient
  unidimensional slam.
\newblock In {\em 2019 IEEE/RSJ International Conference on Intelligent Robots
  and Systems (IROS)}, pages 4176--4181. IEEE, 2019.

\bibitem{taunyazov20event}
T.~Taunyazoz, W.~Sng, H.~H. See, B.~Lim, J.~Kuan, A.~F. Ansari, B.~Tee, and
  H.~Soh.
\newblock Event-driven visual-tactile sensing and learning for robots.
\newblock In {\em Proceedings of Robotics: Science and Systems}, July 2020.

\bibitem{thakur2018large}
C.~S. Thakur, J.~L. Molin, G.~Cauwenberghs, G.~Indiveri, K.~Kumar, N.~Qiao,
  J.~Schemmel, R.~Wang, E.~Chicca, J.~Olson~Hasler, et~al.
\newblock Large-scale neuromorphic spiking array processors: A quest to mimic
  the brain.
\newblock {\em Frontiers in neuroscience}, 12:891, 2018.

\bibitem{todorov2004optimality}
E.~Todorov.
\newblock Optimality principles in sensorimotor control.
\newblock {\em Nature neuroscience}, 7(9):907--915, 2004.

\bibitem{tsodyks1998neural}
M.~Tsodyks, K.~Pawelzik, and H.~Markram.
\newblock Neural networks with dynamic synapses.
\newblock {\em Neural computation}, 10(4):821--835, 1998.

\bibitem{uno1989formation}
Y.~Uno, M.~Kawato, and R.~Suzuki.
\newblock Formation and control of optimal trajectory in human multijoint arm
  movement.
\newblock {\em Biological cybernetics}, 61(2):89--101, 1989.

\bibitem{volinski2022data}
A.~Volinski, Y.~Zaidel, A.~Shalumov, T.~DeWolf, L.~Supic, and E.~E. Tsur.
\newblock Data-driven artificial and spiking neural networks for inverse
  kinematics in neurorobotics.
\newblock {\em Patterns}, 3(1):100391, 2022.

\bibitem{wang2019smooth}
H.~Wang, H.~Wang, J.~Huang, B.~Zhao, and L.~Quan.
\newblock Smooth point-to-point trajectory planning for industrial robots with
  kinematical constraints based on high-order polynomial curve.
\newblock {\em Mechanism and Machine Theory}, 139:284--293, 2019.

\bibitem{whishaw1996endpoint}
I.~Q. Whishaw.
\newblock An endpoint, descriptive, and kinematic comparison of skilled
  reaching in mice (mus musculus) with rats (rattus norvegicus).
\newblock {\em Behavioural Brain Research}, 78(2):101--111, 1996.

\bibitem{windhorst1996role}
U.~Windhorst.
\newblock On the role of recurrent inhibitory feedback in motor control.
\newblock {\em Progress in neurobiology}, 49(6):517--587, 1996.

\bibitem{winters2012biomechanics}
J.~M. Winters and P.~E. Crago.
\newblock {\em Biomechanics and neural control of posture and movement}.
\newblock Springer Science \& Business Media, 2012.

\bibitem{wu1997presynaptic}
L.-G. Wu and P.~Saggau.
\newblock Presynaptic inhibition of elicited neurotransmitter release.
\newblock {\em Trends in neurosciences}, 20(5):204--212, 1997.

\bibitem{xiao2012smooth}
Y.~Xiao, Z.~Du, and W.~Dong.
\newblock Smooth and near time-optimal trajectory planning of industrial robots
  for online applications.
\newblock {\em Industrial Robot: An International Journal}, 2012.

\bibitem{yin2003motion}
K.~Yin, M.~B. Cline, and D.~K. Pai.
\newblock Motion perturbation based on simple neuromotor control models.
\newblock In {\em 11th Pacific Conference onComputer Graphics and Applications,
  2003. Proceedings.}, pages 445--449. IEEE, 2003.

\bibitem{zaidel2021neuromorphic}
Y.~Zaidel, A.~Shalumov, A.~Volinski, L.~Supic, and E.~E. Tsur.
\newblock Neuromorphic nef-based inverse kinematics and pid control.
\newblock {\em Frontiers in Neurorobotics}, 15, 2021.

\bibitem{zhao2022learning}
J.~Zhao, M.~Monforte, G.~Indiveri, C.~Bartolozzi, and E.~Donati.
\newblock Learning inverse kinematics using neural computational primitives on
  neuromorphic hardware.
\newblock 2022.

\bibitem{zhao2020closed}
J.~Zhao, N.~Risi, M.~Monforte, C.~Bartolozzi, G.~Indiveri, and E.~Donati.
\newblock Closed-loop spiking control on a neuromorphic processor implemented
  on the icub.
\newblock {\em IEEE Journal on Emerging and Selected Topics in Circuits and
  Systems}, 10(4):546--556, 2020.

\end{thebibliography}
